\pdfoutput=1

\documentclass[11pt]{article}

\usepackage[final]{acl}

\usepackage{times}
\usepackage{latexsym}
\usepackage{enumitem}
\usepackage{amssymb}
\usepackage{mdframed}
\usepackage{fancyvrb}
\usepackage[T1]{fontenc}

\usepackage[utf8]{inputenc}

\usepackage{microtype}

\usepackage{inconsolata}
\usepackage{amsmath}
\usepackage{cleveref}
\usepackage{graphicx}
\usepackage{subcaption}
\usepackage{booktabs}
\usepackage{float}

\title{Turning Conversations into Workflows:  A Framework to Extract and Evaluate Dialog Workflows for Service AI Agents}

\author{\quad Prafulla Kumar Choubey \quad Xiangyu Peng \quad Shilpa Bhagavath \\
 {\bf Caiming Xiong  \quad Shiva Kumar Pentyala \quad Chien-Sheng Wu}\\
Salesforce AI Research \\
\small{
   \textbf{Correspondence:} \href{mailto:pchoubey@salesforce.com}{pchoubey@salesforce.com}} \\
}

\begin{document}
\maketitle
\begin{abstract}
Automated service agents require well-structured workflows to provide consistent and accurate responses to customer queries. However, these workflows are often undocumented, and their automatic extraction from conversations remains unexplored. In this work, we present a novel framework for extracting and evaluating dialog workflows from historical interactions. Our extraction process consists of two key stages: (1) a retrieval step to select relevant conversations based on key procedural elements, and (2) a structured workflow generation process using a question-answer-based chain-of-thought (QA-CoT) prompting. To comprehensively assess the quality of extracted workflows, we introduce an automated agent and customer bots simulation framework that measures their effectiveness in resolving customer issues. 
Extensive experiments on the ABCD and SynthABCD datasets demonstrate that our QA-CoT technique improves workflow extraction by 12.16\% in average macro accuracy over the baseline.
Moreover, our evaluation method closely aligns with human assessments, providing a reliable and scalable framework for future research.

\end{abstract}

\section{Introduction}

LLMs are changing customer service by helping automated bots respond more accurately and consistently to customer inquiries. These bots use well-structured workflows to guide their interactions and ensure efficient issue resolution and adherence to company policies~\cite{brown2023automate}. However, high-quality workflows are scarce, making it challenging to maintain agent consistency and quality.

\begin{figure}[tb]
    \centering
    \includegraphics[width=0.92\linewidth]{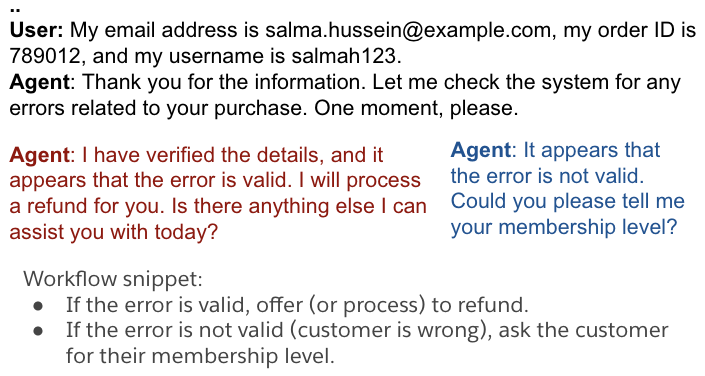}
    \caption{An example showing the derivation of a workflow from historical conversations. Full workflow is shown in Figs. \ref{fig:retriever_1_2} and \ref{fig:eval_1_2}.}
    \label{fig:example}
\end{figure}

Traditional workflow creation methods often rely on manual efforts~\cite{BOHUS2009332,mosig2020star,mehri-eskenazi-2021-schema,chen-etal-2021-action,zhao-etal-2023-anytod,robino2025conversationroutinespromptengineering}, which can quickly become outdated and fail to capture the full nuances of real-world interactions. Notably, common procedural knowledge is embedded within past customer-agent interactions. For instance, Fig. \ref{fig:example} presents a segment of a workflow derived based on two conversations representing different customer scenarios. Yet extracting and structuring this information is unexplored. Automating the extraction of workflows from conversations offers a promising solution by transforming implicit knowledge into structured guidelines for both human agents and AI systems. 

In this paper, we propose a novel two-step pipeline for automatically extracting dialog workflows from customer-agent interactions. These conversations often contain noise, such as agents combining multiple steps or sequencing them incorrectly. Moreover, customer interactions usually cover only a subset of the established rules, making it difficult to capture a complete workflow.
To tackle these challenges, our pipeline first enhances the retrieval process by identifying key procedural components (Fig.~\ref{procedural-elements})—such as intent, slot-values, and resolution steps—and selecting the most relevant conversations while filtering out non-compliant ones (Fig.~\ref{fig:retriever2}). 

In the second step, we introduce a structured QA-based chain-of-thought (QA-CoT) prompting technique to systematically extract workflows from large conversation datasets. Extracting workflows at scale is challenging, as it requires capturing all critical decision points and ensuring their correct representation within the workflow.
Our QA-CoT approach addresses this by simulating an interactive exchange between a Guide and an Implementer agent (Fig.~\ref{qa-cot-example}). The Guide systematically examines key aspects of the workflow, including preconditions, decision points, and step rationales, while the Implementer responds with insights derived from historical conversations. This structured QA process helps the LLM focus on each decision point individually, ensuring that all essential procedural elements are accurately incorporated into the final workflow.

\begin{figure}[]
\tiny
\begin{mdframed}
\begin{verbatim}
"intent": "Customer wants to refund an order before shipping.", 
"slot_values": {
    "full_name": "Chloe Zhang", 
    "username": "czhang94", 
    "email": "czhang94@email.com", 
    "order_id": "1553732700", 
    "refund_method": "original credit card", 
    "shipping_status": "not shipped"
}, 
"resolution_steps": [
    "Agent asked for the customer's full name.", 
    "Agent requested the username, email, and order ID.",
    "Agent asked how the customer would like the refund processed.",
    "Agent inquired about the account ID.", 
    "Agent asked about the shipping status of the order."
]
\end{verbatim}
\end{mdframed}
\caption{An example of procedural elements extracted from a conversation by the GPT-4o mini LLM.} 
\label{procedural-elements}
\end{figure}

Another key contribution of our work is a robust multi-step evaluation pipeline for dialog workflows, featuring an end-to-end method that aligns closely with human assessments. We simulate interactions between a service agent bot (equipped with the predicted workflow and system data) and a customer bot (provided intent and user-specific details). After the conversation ends, we evaluate task success by comparing the outcome achieved using the predicted workflow with the expected outcome derived from the ground truth workflow (see Fig.~\ref{fig:eval2} for an example). 


We evaluate our proposed QA-CoT method against common prompting strategies using the ABCD~\cite{chen-etal-2021-action} and an LLM-generated SynthABCD datasets. ABCD originally grounds conversations on predefined dialog workflows, but we re-purpose it for extracting workflows from conversations. Additionally, to eliminate human-induced errors in ABCD conversations, we generate SynthABCD using LLMs, ensuring error-free conversations while complying with the same workflows. This allows for a controlled and reliable evaluation under an idealized setting.

Our QA-CoT outperforms other prompting methods across different LLMs including GPT-4o, Opus, Sonnet, and Gemini. For instance, on the ABCD dataset, it achieves an 11.81\% gain over the baseline with GPT-4o.
Furthermore, incorporating explicit QA-CoT reasoning also improves the performance of the state-of-the-art reasoning LLMs, including OpenAI o1, o1-mini, o3-mini, as well as DeepSeek-R1 \cite{guo2025deepseek} on both datasets.



\begin{figure*}[t]
    \centering
    \begin{subfigure}[b]{0.49\textwidth}
        \centering
        \includegraphics[width=\textwidth]{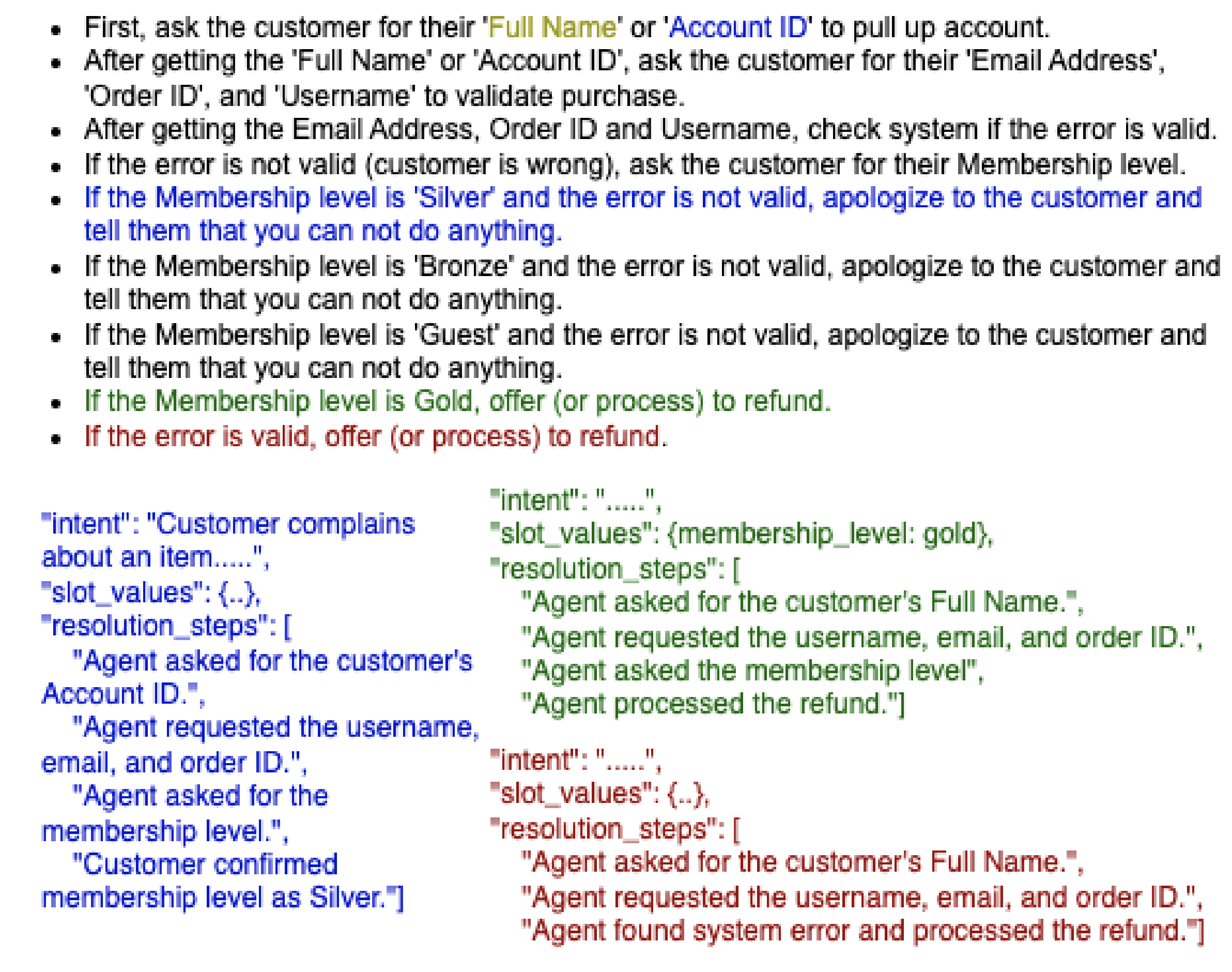}
        \caption{}
        \label{fig:retriever1}
    \end{subfigure}
    \hfill
    \begin{subfigure}[b]{0.49\textwidth}
        \centering
        \includegraphics[width=\textwidth]{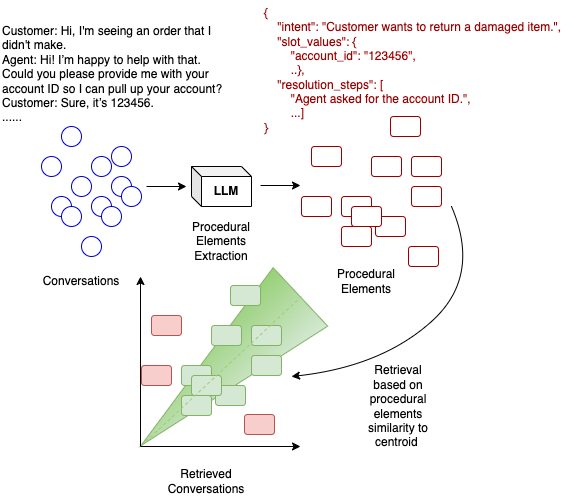}
        \caption{}
        \label{fig:retriever2}
    \end{subfigure}
    \caption{(a) An example workflow with procedural elements for 3 of the 10 possible distinct sub-flows. Sub-flows and their procedural elements are color-coded to match. (b) Steps in our proposed procedural element-based retriever. An example of complete procedural elements extracted from a conversation is shown in Fig. \ref{procedural-elements}.}
    \label{fig:retriever_1_2}
\end{figure*}

\begin{figure*}[!tb]
\tiny 
\begin{mdframed}
\begin{minipage}{\linewidth} 
\begin{Verbatim}[commandchars=\\\{\}]
..
\textbf{Guide:} What additional details are required to validate the customer's purchase?
\textbf{Implementer:} The agent must ask for the username, email address, and order ID to validate the purchase.
\textbf{Guide:} How does the customer's membership level impact the return process?
\textbf{Implementer:} The customer's membership level determines the return policy duration: 
                    - \textbf{Guest membership} requires returns within \textbf{30 days} with receipt.
                    - \textbf{Bronze membership} allows returns within \textbf{90 days} or with original packaging.
                    - \textbf{Silver membership} permits returns \textbf{within six months}.
                    - Gold members have unlimited time for returns.
\textbf{Guide:} What should the agent do if the return is past the allowable period for the customer's membership level?
\textbf{Implementer:} If the return period has passed, the agent should check if the customer has the original receipt and packaging. If not, they politely 
inform the customer that a return cannot be processed.
..
\end{Verbatim}
\end{minipage}
\end{mdframed}
\caption{A snippet of the QA chain-of-thought generated by the GPT-4o model for the $return\_color$ intent using conversations from the ABCD data. The extracted QA pairs highlight key preconditions based on membership level.}
\label{qa-cot-example}
\end{figure*}


\section{Related Work}
Dialog workflows are a specialized form of procedural knowledge. While workflow extraction has received little attention, automatic procedural extraction has been widely studied, primarily focusing on ``how-to'' documents \citep{10.1145/2187980.2188194,maeta-etal-2015-framework,Chu2017DistillingTK,Park2018LearningPF} and instructional videos \citep{ushiku-etal-2017-procedural,10.5555/3504035.3504965,xu-etal-2020-benchmark}.
These works typically model linear sequences of explicitly stated actions, aiming to either predict procedural steps or generate summaries of task execution \citep{Koupaee2018WikiHowAL}.
In contrast, our work tackles dialog workflows, where actions are often implicit and depend on previous steps, user inputs, and system responses. This introduces decision-dependent variability, making extraction significantly more challenging than predicting fixed procedural sequences.

Specific to dialog systems, there has been extensive research on studying structures in task-oriented dialogs~\cite{Jurafsky1997SwitchboardSS,Chotimongkol2008LearningTS,shi2019unsuperviseddialogstructurelearning,xu2020discoveringdialogstructuregraph,chen2021dsbertunsuperviseddialoguestructurelearning,Nath2021TSCAND,wang2021modellinghierarchicalstructuredialogue,rony2022dialokgknowledgestructureawaretaskoriented,lu2022unsupervisedlearninghierarchicalconversation,qiu2022structureextractiontaskorienteddialogues,Yin_2023,pryor2024usingdomainknowledgeguide,burdisso2024dialog2flow}, focusing on how dialogs evolve using dialog acts, intent-slot pairs, or turn-level dependencies. Our work parallels workflow discovery \cite{hattami2023workflow, raimondo2023improving,min2023workflowguided}, which aims to predict the optimal next dialog action from the conversation's current state and all available actions. However, unlike this, we focus on extracting global workflows applicable across conversations for a specific intent. This increases the complexity of the task, as the model must filter out noisy actions, and consolidate multiple potential actions sequences from different conversations. Additionally, we also propose a new QA-CoT prompting method for workflow extraction and introduce a robust  end-to-end evaluation framework to assess the accuracy of the extracted workflows.
 
\section{Dialog Workflow Extraction }
\begin{figure*}[t]
    \centering
    \begin{subfigure}[b]{0.49\textwidth}
        \centering
        \includegraphics[width=\textwidth]{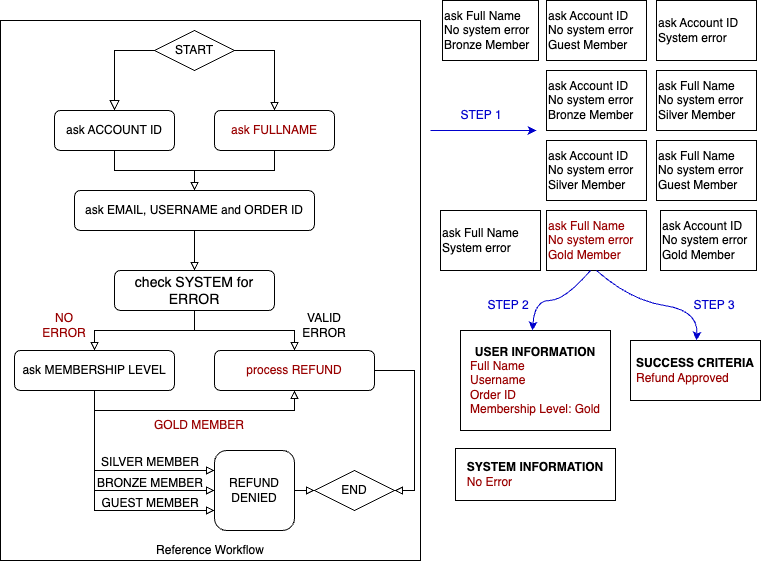}
        \caption{}
        \label{fig:eval1}
    \end{subfigure}
    \hfill
    \begin{subfigure}[b]{0.49\textwidth}
        \centering
        \includegraphics[width=\textwidth]{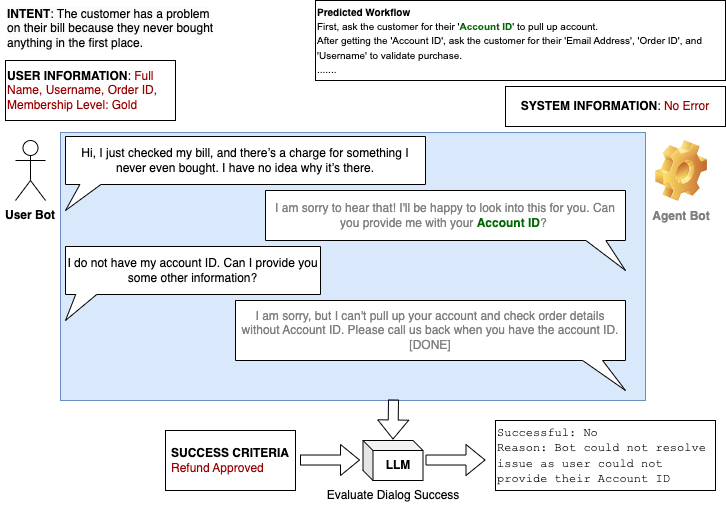}
        \caption{}
        \label{fig:eval2}
    \end{subfigure}
    \caption{(a) Flowchart of the workflow in Fig. \ref{fig:retriever1}, illustrating 10 possible customer scenarios [Step 1, E2E pipeline], along with an example of user information, system information, and the success criteria for one scenario [Steps 2 and 3]. (b) Simulation of the user (system) bot based on intent and user information (system information and the predicted workflow) [Step 4], followed by a final evaluation of dialogue success [Step 5].}
    \label{fig:eval_1_2}
\end{figure*}

The dialog workflow extraction task aims to identify and organize all essential steps required to resolve customer issues across various scenarios into structured guidelines. For instance, Fig. \ref{fig:retriever1} illustrate a workflow for customer complaints about a bill they never purchased, while Fig. \ref{fig:eval1} presents the corresponding state machine diagram, highlighting different sub-flows. It covers scenarios such as assisting gold, silver, guest, or bronze members; addressing cases with or without system errors; and accommodating customers who provide either their full name or account details.
In this example, we aim to generate a workflow that covers all variables, including membership levels, system errors, and available customer information.

We focus on two key aspects to improve dialog workflows extraction: retrieval, which involves selecting the most relevant historical conversations for the task, and extraction, which uses a structured QA-CoT prompting approach to derive workflows from the selected conversations.

\subsection{Procedural Element-based Retrieval}
Our goal is to extract all valid step sequences for resolving customer issues, ensuring comprehensive scenario coverage while filtering out incorrect or noisy conversations. Since our focus is on the underlying process instead of surface-level dialogs, we propose a two-step selection strategy.
First, we extract key procedural elements—such as intent, relevant slot values, and the resolution steps taken by the agent—from each conversation using GPT-4o-mini (example in Figs.~\ref{procedural-elements} and ~\ref{fig:retriever1}, prompt in Fig. \ref{prompt:conv-selection-procedure}).
This ensures selection is based on procedural content rather than surface-level dialog.

Next, we embed each conversation using its extracted key procedural elements with the OpenAI text-embedding-3-small model, then we cluster and compute the centroid for each intent. We then select the top-K conversations closest to each centroid using cosine similarity, ensuring the chosen conversations represent the most commonly followed resolution steps (Fig.~\ref{fig:retriever2}). By focusing on key procedural elements rather than full conversations, we retain meaningful process information while filtering out noise. 
As we demonstrate in $\S$\ref{result:retriever}, directly clustering full conversation embeddings captures extraneous details and conversational variability, leading to significantly lower performance compared to procedural element embeddings.

While this approach effectively captures frequent resolution steps, it may reduce diversity. Given the inherent noise in conversations, distinguishing between true procedural variations and spurious deviations remains a challenge for LLMs. Our strategy prioritizes common procedural patterns while minimizing noise, enabling more reliable workflow extraction.

\subsection{QA-CoT for Workflows Extraction}

A naive approach to extracting workflows is to use simple prompting (Fig.\ref{prompt:procedural-extraction-basic}) with retrieved conversations. However, this often results in workflows that lack fine-grained details. For instance, in the example workflow shown in Fig.\ref{fig:retriever1}, the generated workflow might rigidly follow a sequence by checking for errors first and then asking the membership level even when a system error occurs.

To address these limitations, we propose QA-CoT, a structured framework inspired by multi-agent debate and collaboration systems~\cite{du2023improvingfactualityreasoninglanguage,zhang2024chainagentslargelanguage,musumeci2024llmbasedmultiagentgeneration,hu2025debatetowritepersonadrivenmultiagentframework}. In QA-CoT, a Guide-Implementer interaction generates QA pairs (example in Fig.~\ref{qa-cot-example}, prompt in Fig.~\ref{prompt:qa-cot}) to capture fine-grained procedural knowledge. 
The Guide formulates targeted questions based on past conversations, focusing on clarifying preconditions, decision points, and logical dependencies. For instance, in the example workflow shown in Fig.~\ref{fig:retriever1}, the Guide would explicitly ask: ``What is the next step if there is a system error?'' and ``What is the next step if there is no system error?'' The Implementer, using the same past conversations, provides detailed responses, such as ``asking for the membership level'' or ``processing a refund'' based on the scenario. To improve workflow extraction, we augment historical conversations with the Guide-Implementer exchange and prompt the LLM (Fig.~\ref{prompt:procedural-extraction}) to generate workflows that incorporate fine-grained procedural details.

We experimented with two approaches for simulating this interaction. The first is multi-turn prompting, where the Guide and Implementer take turns. The Guide asks a question, and the Implementer provides a response. This process continues until the Guide has no further clarification questions or a maximum of 25 turns is reached. The second approach generates the entire exchange in a single pass. Empirically, the single-pass method yields superior results with GPT-4o (results in Table~\ref{tab:multi-step-vs-single-step-qa}), while also being computationally efficient. Given our large-scale experiments across 8 LLMs and two datasets, we adopt the single-pass strategy.

\section{Dialog Workflow Evaluation}
Evaluating workflow requires careful attention to the relationships and dependencies between constituent steps to ensure the predicted workflow aligns with the expected decision-making process. We encountered significant challenges in performing this evaluation at the workflow level, primarily due to the complex dependencies between steps that are difficult to assess consistently. 
In an initial study, we asked human annotators to evaluate whether each step in the ground-truth workflows was accurately represented in the predicted workflows, using a set of 10 LLM-extracted workflows. The results showed moderate inter-annotator agreement, with a Cohen's $\kappa$ score of 0.45, highlighting the difficulties even humans face in identifying nuanced procedural differences.

To address these challenges, we propose an end-to-end (E2E, Fig.\ref{fig:eval_1_2}) evaluation framework that works in five stages\footnote{We also explores alternative evaluation strategies such as QA-based evaluation~\cite{fabbri2021qafacteval}. These strategies proved less effective than our proposed E2E approach and are discussed in detail in Appendix~\ref{sec:alternative-evaluation}.}:

\begin{itemize}[leftmargin=*]
    \item \textit{Decomposing Workflows into Sub-flows}: We first decompose each workflow into its constituent sub-flows. As shown in Fig. \ref{fig:eval1} (Step 1), this process involves extracting all 10 possible paths from START to END, which result from variations in membership levels, the type of user input (full name or account ID), and the occurrence of system errors. This approach ensures comprehensive coverage of all decision branches in our evaluation (prompt shown in Fig. \ref{prompt:synthabcd-e2e-step1}).
    \item \textit{Generating User and Agent Bot Information}: Next, we map each sub-flow to the relevant user and system information required for bot simulation. In the subflow highlighted in red in Fig. \ref{fig:eval1} (Step 2), the user bot should provide only their full name, username, order ID, and membership level (e.g., silver), while the agent bot has access to internal systems that confirm no error occurred on the company’s end (prompt shown in Fig. \ref{prompt:e2e-step2}).
    \item \textit{Defining Success Criteria}: We further link each sub-flow to a success criterion, which represents the expected outcome of the conversation if the agent follows the ground-truth workflow. In the sub-flow highlighted in red in Fig. \ref{fig:eval1} (Step 3), the agent is expected to approve the user's refund request (prompt shown in Fig. \ref{prompt:e2e-step2}).
    \item \textit{Dialog Simulation}: Next, we simulate interactions between a customer bot and a service agent bot, where the customer bot conveys intent and provides user-specific details, while the agent bot utilizes system data and acts according to the predicted workflow. The customer bot responds strictly to explicit agent's requests or seeks alternatives when unable to provide the requested information. Meanwhile, the agent bot executes the predicted workflow by requesting data (e.g., asking for a username), performing actions (e.g., issuing refunds, checking system information), or terminating the conversation if no further steps are possible. The interaction ends when the agent either completes the workflow successfully or cannot proceed due to missing information (Fig. \ref{fig:eval2}, prompts in Figs. \ref{prompt:customer-bot-e2e} and \ref{prompt:service-bot-e2e}).
    \item \textit{Success Evaluation}: Once the conversation ends, we evaluate whether the simulated interaction meets the success criteria. For instance, in Fig. \ref{fig:eval2}, the interaction fails because the user bot was unable to provide the account ID (prompt shown in Fig. \ref{prompt:e2e-success}).
\end{itemize}

We evaluated the accuracy of our E2E simulation pipeline and found that the overall accuracy across steps 1 to 4 was 94.81\%. Moreover, step 5 exhibited strong alignment with human evaluation, achieving a Cohen’s Kappa score of 0.92. These results suggest that our E2E evaluation closely mirrors human judgments (for details, see $\S$\ref{appendix:e2e-human}).

\section{Experimental Settings}
\subsection{Dataset}
We use two datasets in our evaluations: the human-developed ABCD and the LLM-synthesized SynthABCD. The ABCD dataset is inherently noisy ($\S$~\ref{sec:data-compliance}), as conversations were generated through human interactions between a customer and a support agent. This process introduced inconsistencies, such as customers providing information before being prompted and agents deviating from workflows by reordering, skipping, or merging steps. While these variations reflect real-world interactions, they complicate workflow extraction. To address these challenges, we also create SynthABCD, a synthetic dataset derived from ABCD workflows using an LLM. SynthABCD reduces noise by strictly following predefined workflows, enabling a more accurate evaluation of LLMs' capabilities.

In the first step of the SynthABCD synthesis, we use the same step 1 from E2E evaluation framework to generate all possible sub-flows. 
Next, we pair each subflow with the corresponding workflow to guide the LLM in simulating several conversations between a customer and an agent for each subflow.
Additional data details, such as the intents used for ABCD and SynthABCD and human evaluation for SynthABCD, are discussed in $\S$\ref{sec:appendix-data}. 

\subsection{Conversations Retrievers}
We evaluate our procedural element similarity-based retrieval strategy (Proc-Sim) against three baselines: random selection, procedural element diversity-based retrieval (Proc-Div), and conversation similarity-based retrieval (Conv-Sim). 
For Conv-Sim, we use the same approach as Proc-Sim, except that we use full conversation embeddings rather than procedural element embeddings.
For Proc-Div, we want to enrich diversity of selected conversations, by first removing noisy conversations by excluding those in the lowest 10th percentile based on cosine similarity to the centroid of procedural element embeddings. Then, from the remaining conversations, we select one conversation furthest from the centroid as the initial candidate. Then, we iteratively select additional conversations that are maximally distant from the centroid of the procedural elements of the already selected conversations~\cite{GONZALEZ1985293}. 

\begin{figure*}[ht!]
    \centering
    \begin{subfigure}[b]{0.49\textwidth}
        \centering
        \includegraphics[width=\textwidth]{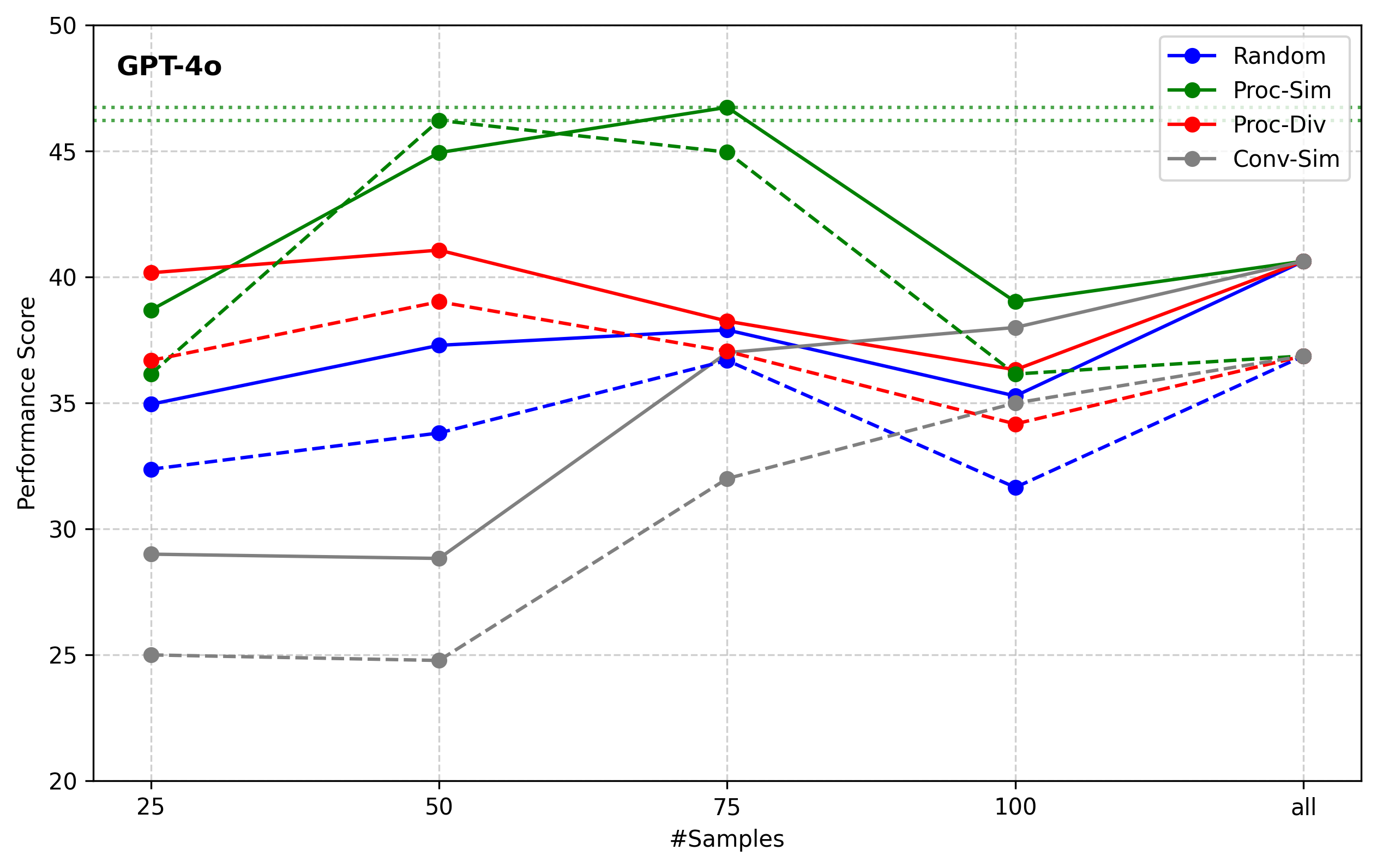}
        \label{fig:f1}
    \end{subfigure}
    \hfill
    \begin{subfigure}[b]{0.49\textwidth}
        \centering
        \includegraphics[width=\textwidth]{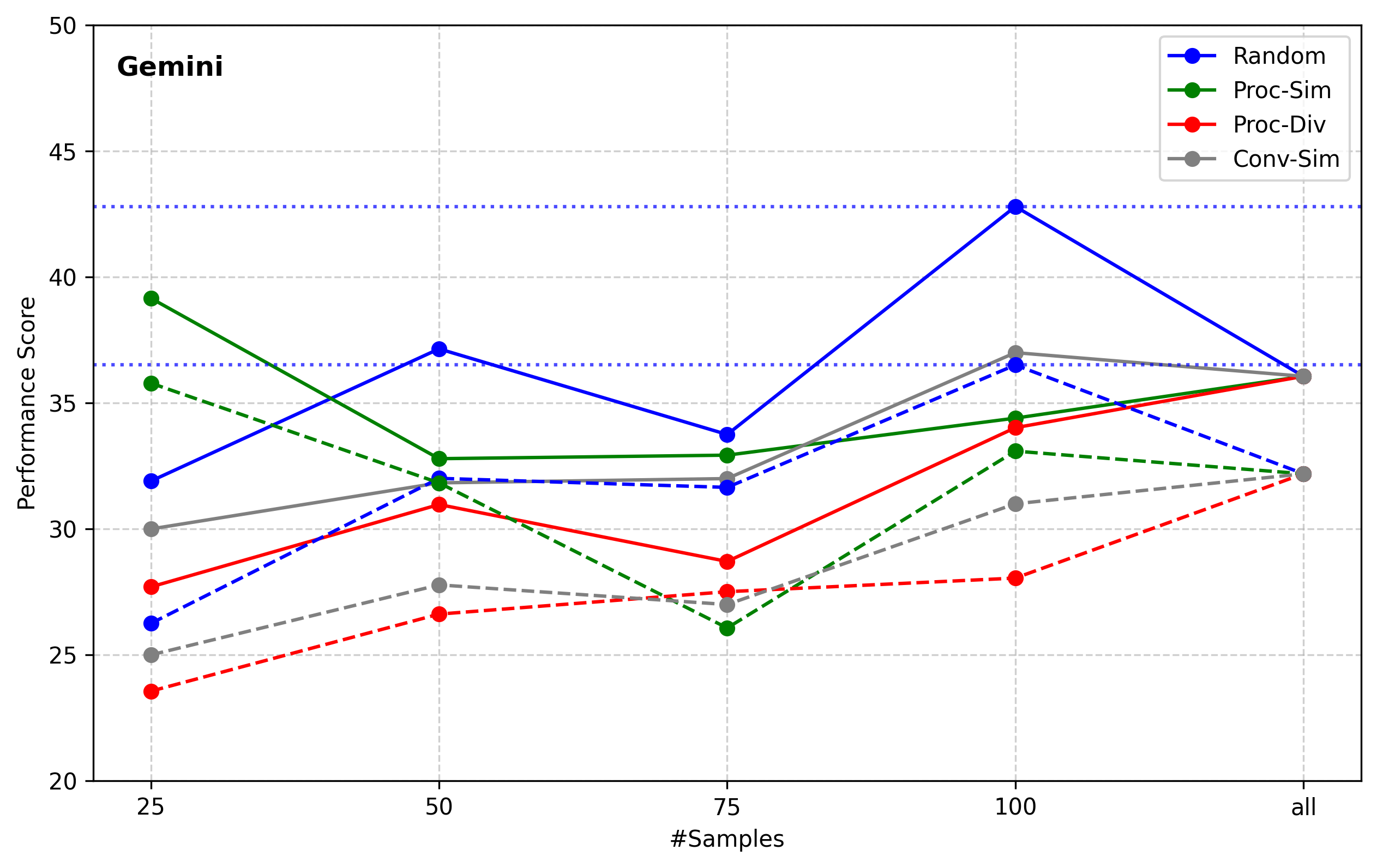}
        \label{fig:f2}
    \end{subfigure}

    \vspace{1em} 

    \begin{subfigure}[b]{0.49\textwidth}
        \centering
        \includegraphics[width=\textwidth]{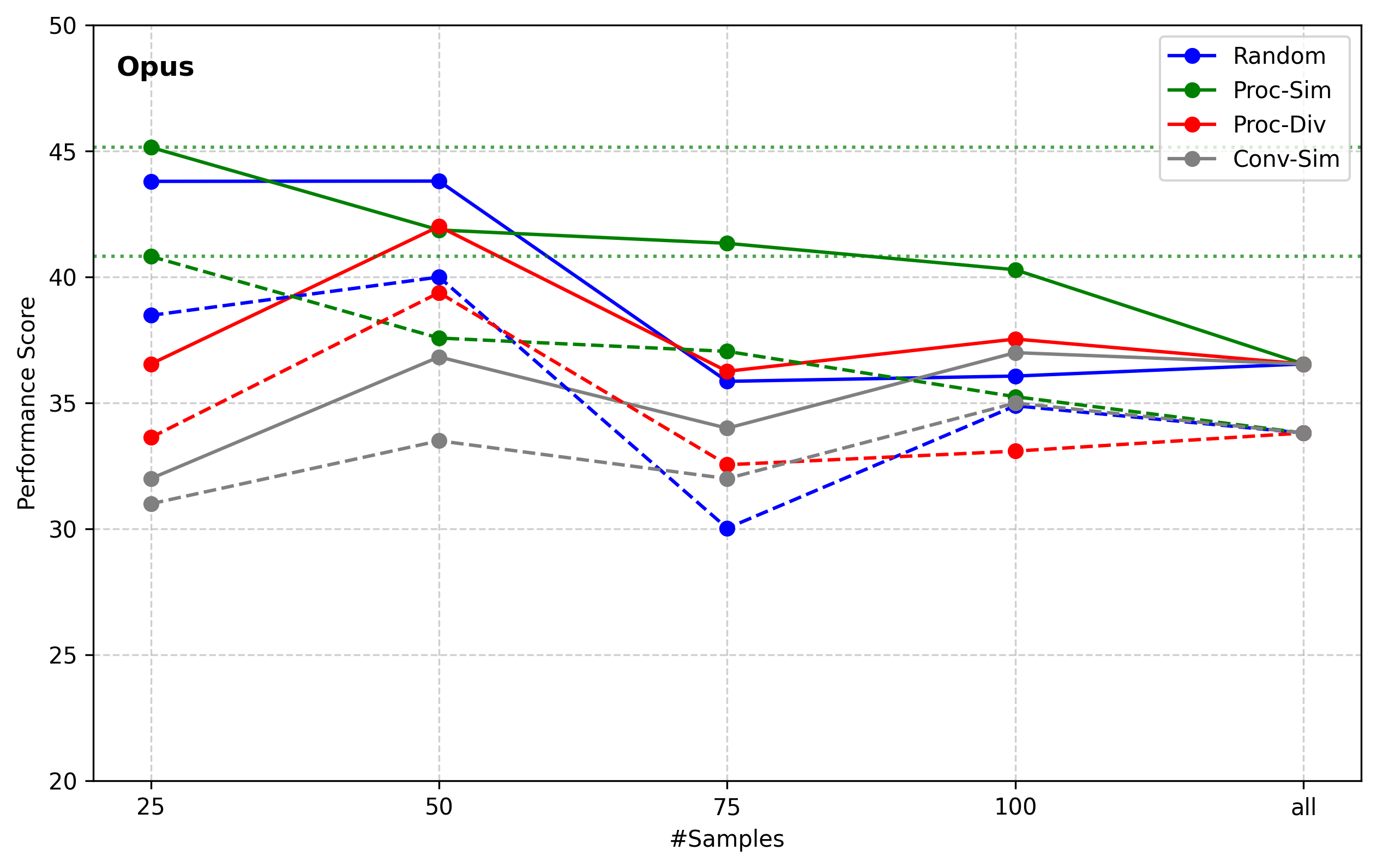}
        \label{fig:f3}
    \end{subfigure}
    \hfill
    \begin{subfigure}[b]{0.49\textwidth}
        \centering
        \includegraphics[width=\textwidth]{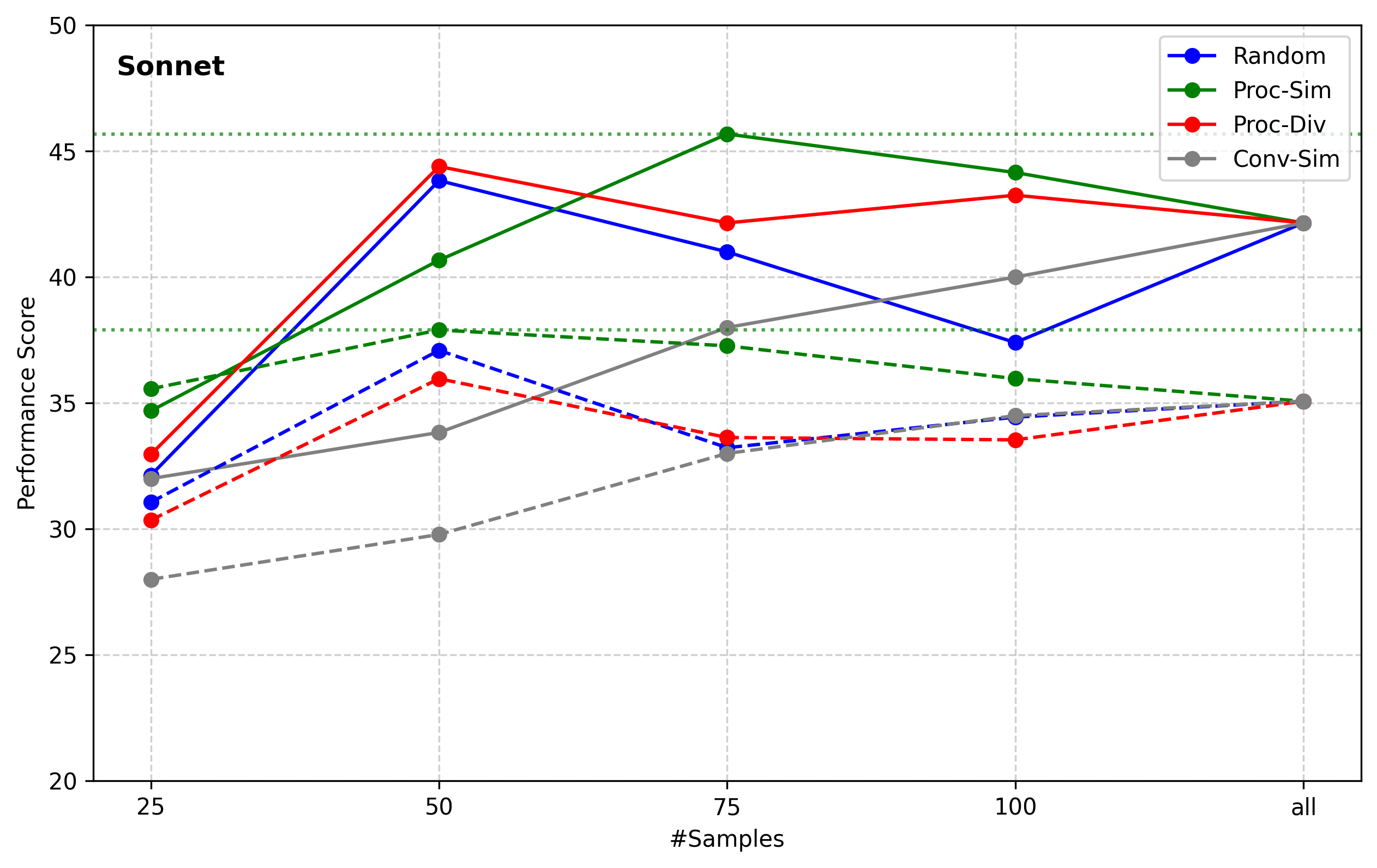}
        \label{fig:f4}
    \end{subfigure}
    
    \caption{Performance comparison of conversation selection strategies (Proc-Sim, Conv-Sim, Random, and Proc-Div) with varying numbers of selected conversations. Results are shown for four LLMs: gpt-4o, gemini-1.5-pro, opus-3, and sonnet-3.5, using E2E macro (solid lines) and micro (dashed lines) accuracies.
    }
    \label{fig:retriever}
\end{figure*}

\subsection{Workflows Generation Method}
We compare the performance of our proposed QA-CoT model against several prompting techniques. 
\begin{itemize}[leftmargin=*]
    \item Basic: We use a single-step prompt over selected conversations to generate workflows directly (prompt in Fig.~\ref{prompt:procedural-extraction-basic}). 
    
    \item Reflection~\cite{shinn2023reflexionlanguageagentsverbal}: We build upon Basic by identifying and addressing coverage gaps in the workflow. After generating the initial workflow using basic prompting, a reflection prompt is applied to identify missing steps or inconsistencies, which are then refined to improve the workflow iteratively (prompts in Figs.~\ref{prompt:procedural-extraction-reflect} and~\ref{prompt:procedural-extraction-reflect-generate}).

    \item Plan~\cite{wang-etal-2023-plan}: This approach first prompts the LLM to devise a structured plan for extracting workflows given the conversations and then applies a second LLM prompt to execute the plan and generate workflows from the same set of conversations (prompts in Figs.~\ref{prompt:procedural-extraction-plan} and~\ref{prompt:procedural-extraction-plan-generating}).

    \item Ensemble: We start by generating four workflows using basic prompting, each based on a different random order of conversations. The LLM then evaluates these workflows, checking step-by-step correctness based on cross-consistency among them, and produces a final consolidated workflow by selecting the most consistent steps across all versions~\cite{chen2023universalselfconsistencylargelanguage} (prompt in Fig.~\ref{prompt:procedural-extraction-ensemble}).

    \item QA-CoT+Reflect: We apply a reflection prompt to the QA chain-of-thought generated through the Guide-Implementer interaction, identifying and refining missing questions before generating the final workflow (prompt in Fig.~\ref{prompt:procedural-extraction-qa-reflect}).
\end{itemize}

For consistency, all numbers are averaged using two workflows per intent by randomizing the order of conversations in all experiments.  We evaluate performance using macro accuracy (arithmetic average of the percentage of sub-flows correct per intent), micro accuracy (percentage of total subflows correct across all intents), and \#utt, which represents the average number of utterances in simulated conversations.


\section{Results and Analysis}

\begin{table*}
\centering
\small
\begin{tabular}{l|rrr|rrr|rrr|rrr} \hline
& \multicolumn{3}{c|}{GPT-4o} & \multicolumn{3}{c|}{gemini}  & \multicolumn{3}{c|}{opus}  & \multicolumn{3}{c}{sonnet} \\ \hline
Method    & Macro  & Micro  &  \#utt   & Macro  & Micro  &  \#utt  & Macro  & Micro & \#utt    & Macro  & Micro  &   \#utt       \\ \hline
 \multicolumn{13}{c}{ABCD} \\ \hline
Basic & 46.74 &	44.96 & 8.69 & 42.80	& 36.51	& 7.89 & 45.15 & 40.82 & 9.99 & 45.68 & 41.0 & 8.86 \\
Reflect  & 44.43 & 42.09 & 8.51  & 37.14 & 35.61 & 8.51 & 50.89 & 45.86 & 10.47 & 51.80  & 49.46 & 10.57 \\ 
Plan & 37.29 & 38.30 & 7.71 & 41.0 & 38.12 & 6.75 & 35.77 & 32.91 & 7.90 & 37.56 & 34.71 & 7.39  \\
Ensemble & 46.13 & 45.32 & 8.66 & 38.71 & 38.67 & 8.42 & 41.80 & 34.35 & 8.51 & 46.47 &	42.44 &	9.29 \\ \hline
QA-CoT  &  \textbf{58.55} & \textbf{56.29} & 10.60  & \textbf{48.09} & \textbf{46.40}  & 8.40  & \textbf{52.36} &	\textbf{55.75} &	11.92 &  \textbf{56.28} & \textbf{53.44} & 11.49 \\
 + Reflect &  55.45 & 55.39 & 10.73 & 43.10  & 46.22 & 8.36 & 56.84 & 56.65 & 11.54 & 54.44 &	54.67 &	10.56 \\ 
 \hline
 \multicolumn{13}{c}{SynthABCD} \\ \hline
Basic & 68.91 & 61.95 & 8.71  & 74.73 & 75.65 & 8.59  & 91.43 & 91.52 & 12.29 & 51.23 & 45.43 & 7.49  \\
Reflect  & 57.42	& 52.61	& 8.24  & 61.90  & 60.43 & 9.08  & \textbf{91.85} & \textbf{92.82} & 12.47 & 49.14 &	41.08 &	7.68  \\ 
Plan & 78.40 & 68.91 & 8.55 & 72.77 & 70.65 & 7.97 & 74.85 & 70.21 & 9.07 & 73.02 & 70.65 & 9.32 \\
Ensemble & 84.31 & 81.30 & 9.62 & 81.13 & 79.78 & 10.41 & 78.70 & 73.26 & 9.21 & 55.97 & 45.43 & 7.99 \\ \hline
QA-CoT  & \textbf{86.53} & \textbf{86.08} & 11.96 & \textbf{84.04} & \textbf{81.96} & 9.70 & 89.73 & 90.21 & 12.27 & \textbf{88.38} & \textbf{86.73} & 10.46 \\
 + Reflect & \textbf{87.87} & \textbf{89.13} & 12.28 & 76.59 & 73.91 & 9.89  & 87.12 & 85.21 & 11.83 & 82.59 & 85.65 & 11.08 \\ 
 \hline
\end{tabular}
\caption{Performance comparison of different workflow extraction systems.} \label{tab:generator-e2e}
\end{table*}

\begin{table}[t]
\centering
\small
\begin{tabular}{lccc}
\hline
 & Macro  & Micro  &  \#utt \\ \hline
Ground-truth Workflow    & 96.93 & 96.97 &  10.58  \\ \hline
\end{tabular}
\caption{Performance when ground-truth workflows are used to simulate the system bot.}
\label{tab:evaluation-gold-e2e}
\end{table}

\subsection{Conversations Retrieval} \label{result:retriever}
For each conversation retrieval approach, we select subsets containing 25, 50, 75, 100, and all conversations and prompt four LLMs (GPT-4o, gemini-1.5-pro, opus-3, and sonnet-3.5) to extract workflows. The performance of each strategy is evaluated using arithmetic macro accuracy (solid lines) and micro accuracy (dashed lines), as shown in Fig. \ref{fig:retriever}. 

Our results show that \textbf{selecting a subset of conversations consistently leads to better performance than using all available conversations}, highlighting that simply applying any LLM to the entire dataset is not the most effective approach. GPT-4o and Sonnet-3.5 achieve peak performance with 75 conversations, Opus-3 peaks at 25, and Gemini-1.5-Pro performs best with 100. 

Among different retrieval strategies, \textbf{Proc-Sim is the most effective for three of the four LLMs, and it is better than Conv-Sim}, suggesting that relying on conversations is less effective than explicit procedural elements. 
Gemini-1.5-Pro is the only model that achieves the best performance under the random selection strategy, implying that it is less sensitive to conversation noise. 
In contrast, Proc-Div consistently yields worse results, indicating that prioritizing diversity introduces noise from real-world conversations. With a more diverse but noisy set of conversations, the model may miss the correct procedure and instead incorporate outlier or incorrect steps, leading to reduced performance.


To systematically evaluate different prompting techniques and LLMs for workflow extraction, we adopt the best-performing retrieval strategy and the number of conversations for each LLM{ separately (e.g., GPT-4o we use Proc-Sim with 75 samples). We use macro-accuracy for measuring performance, which accounts for differences in workflow complexity by treating each workflow equally, rather than skewing results toward those with more sub-flows.

\subsection{Workflows Generation}
As shown in Table \ref{tab:generator-e2e}, all baselines—Reflection, Plan, and Ensemble—fail to consistently outperform basic prompting across LLMs. 
Notably, Reflection and Ensemble incorporate additional refinement steps to improve workflow extraction; however, these refinements do not consistently enhance performance and may introduce noise. Similarly, Reflection prompting on QA-CoT fails to improve performance, reinforcing that refinement steps may not always help workflow extraction.

On the other hand, \textbf{QA-CoT outperforms almost all baselines on both ABCD and SynthABCD datasets}, showing that augmented context with Guide and Implementer agents is helpful for workflow generation. On average, QA-CoT improves macro accuracy by 8.73\% on ABCD and 15.59\% on SynthABCD datasets. The most significant improvement is seen on SynthABCD with the Sonnet model, where QA-CoT boosts macro accuracy by 37.15\%. On ABCD, it achieves the highest gain of 11.81\% for GPT-4o model.
These results highlight QA-CoT's effectiveness in enhancing workflow extraction across different LLMs.
There is one outlier: Opus model for SynthABCD, where reflect prompting performs better. Notably, Opus achieves near-optimal performance on SynthABCD, with a macro accuracy of 91.43\% in this ideal setting, while the upper bound, attained when the agent bot follows the ground-truth workflow, is 96.93\% (Table~\ref{tab:evaluation-gold-e2e}). Unlike predicted workflows, which may introduce errors, the ground-truth workflow guarantees strictly correct actions, representing the theoretical performance ceiling.


We also find that all LLMs exhibit notably higher performance when using SynthABCD, showcasing the strong ability of LLMs to effectively extract workflows from error-free conversations. This further emphasizes the importance of developing accurate conversation selection strategies, which can substantially enhance performance.

Lastly, we observe that ground-truth workflows have an average of 10.58 utterances per conversation (Table~\ref{tab:evaluation-gold-e2e}), whereas model-predicted workflows tend to produce longer interactions despite lower accuracy. For instance, the best-performing predicted workflows, generated by Opus on SynthABCD, have an average of 1.89 utterances higher (\#utt of 12.47) but a macro accuracy 5.08 points lower than the ground-truth workflows. 

Our results also show a positive correlation between macro accuracy and utterance count, suggesting that higher-quality workflows lead to longer conversations. If a predicted workflow splits a correct step into multiple sequential steps, E2E evaluation still treats both as equivalent since the simulated conversation meets the success criteria. To distinguish such cases, we recommend using utterance count (\#utt) as a measure of efficiency, selecting the workflow that achieves the goal in fewer steps.


\begin{table*}
\centering
\small
\begin{tabular}{l|rrr|rrr|rrr|rrr} \hline
& \multicolumn{3}{c|}{o1} & \multicolumn{3}{c|}{o1-mini}  & \multicolumn{3}{c|}{o3-mini}  & \multicolumn{3}{c}{DeepSeek R1} \\ \hline
Method    & Macro  & Micro  &  \#utt   & Macro  & Micro  &  \#utt  & Macro  & Micro & \#utt    & Macro  & Micro  &   \#utt       \\ \hline
 \multicolumn{13}{c}{ABCD} \\ \hline
Basic & 36.72 & \textbf{35.97} & 7.70 & 44.20 & 40.46 & 9.73 & 57.06 & 57.19 & 9.52 & 46.10 & 40.10 & 8.42 \\
QA-CoT  & \textbf{39.0} & 33.09 & 8.50 & \textbf{60.74} & \textbf{64.74} & 12.05 & \textbf{61.20} & \textbf{61.51} & 11.26 &  \textbf{47.24} & \textbf{44.06} & 7.89 \\
 \hline
 \multicolumn{13}{c}{SynthABCD} \\ \hline
Basic & 59.57 & 52.39 & 7.65 & 69.57 & 67.39 & 10.56 & 79.50 & 78.91 & 9.14 & 65.26 & 51.52 & 7.50  \\
QA-CoT  &  \textbf{80.60} & \textbf{76.95} & 9.62 & \textbf{87.16} & \textbf{87.17} & 12.43 & \textbf{89.49} & \textbf{90.21} & 10.31 & \textbf{66.95} & \textbf{55.65} & 8.41 \\
 \hline
\end{tabular}
\caption{End-to-End performance comparison of different reasoning LLMs.} \label{tab:reasoning-e2e}
\end{table*}

\subsection{Performance of Reasoning LLMs}
In Table \ref{tab:reasoning-e2e}, we show performance of four reasoning LLMs—DeepSeek-R1, o1, o1-mini, and o3-mini—on both the ABCD and SynthABCD datasets. We use 75 conversations selected using Proc-Sim retriever.
Our results show that \textbf{simply prompting a reasoning LLM does not consistently improve performance}. Among the evaluated models, only o3-mini demonstrates a significant improvement over non-reasoning LLMs on ABCD, achieving performance levels close to QA-CoT prompting with non-reasoning models.
Furthermore, employing \textbf{QA-CoT with reasoning LLMs consistently improves performance over basic prompting across both datasets}. This finding highlights that explicit QA-based chain-of-thought prompting can even benefit reasoning LLMs in extracting dialog workflows.

To better understand the limitations in performance improvement, we analyze both the thinking process and final outputs of DeepSeek-R1 on the ABCD and SynthABCD datasets (Fig. \ref{deepseek-think-example}). Our analysis reveals that while the thinking tokens capture the general process, they lack fine-grained preconditions. For instance, instead of explicitly detailing the relationship between the number of days and membership type in determining return eligibility, the model generates a generic thinking such as ``determine return eligibility based on purchase date.'' This omission potentially leads to incomplete workflow extraction, as critical decision criteria—such as a 6-month return window for silver members versus a 30-day window for guest members (QA-CoT with GPT-4o includes these fine-grained details as shown in Fig.~\ref{qa-cot-example})—are not captured.

\section{Error Analysis}
We manually reviewed all eight QA-CoT systems to find where they failed to handle all scenarios correctly (example in Fig. \ref{fig:appendix-error}). One common mistake was \textbf{confusing system-available information with details that should be requested from the user}. This issue was especially common in shipping-related tasks, where models wrongly assumed they should check the shipping status in the system instead of asking the customer. o1 and DeepSeek-R1 made this mistake in five intents, while o3-mini was the only model that avoided it. This error did not come from the conversation data but from the LLM’s built-in knowledge, as it appeared in both the ABCD and SynthABCD datasets.
Another common issue was \textbf{models tend to ignore alternative options}. Instead of allowing a choice, they often predicted only one option. For example, when a workflow required asking for either the ``full name or the account ID'', models typically picked just one (e.g., ask ``full name'') instead of keeping both possibilities.
A third major issue was \textbf{failing to follow conditional logic at branching points}. For example, a workflow might require asking for the membership level only if there is no system error. However, models often ignored this condition and predicted a fixed sequence, combining steps incorrectly instead of following the intended logic.

We also observed two other patterns that did not impact performance but affected efficiency. First, models often \textbf{merged multiple steps into a single step}, particularly in the early stages of workflows. For instance, a workflow might specify sequential steps such as verifying the account ID or full name first, followed by email ID, order ID, and username. Instead of keeping these as separate steps, models frequently combined them, which reduces the average number of utterances required to resolve an issue. Second, when the original workflow allowed for multiple conditions to proceed—such as purchase date within 90 days, original receipt, or original packaging—models tended to convert this flexibility into a strict sequence. For example, they predicted checking the purchase date first, then asking about the receipt if the date exceeded 90 days, and so on, rather than allowing any of the conditions to be met.
While this does not reduce accuracy, it negatively impacts efficiency by increasing the number of utterances compared to the ground-truth workflow. 

\section{Conclusion}
We introduced a novel framework for extracting and evaluating dialog workflows from historical conversations. Our proposed QA-CoT prompting outperforms standard prompting strategies, achieving a 12.16\% improvement in average macro-accuracy. Furthermore, the strong alignment of our E2E evaluation with human assessments underscores the robustness and reliability of our evaluation framework. We believe that our work offers a robust foundation for structured workflow extraction and evaluation, and will encourage future research to develop more effective methods. 


\section{Limitations}
Our workflow extraction and evaluation methods are designed for Service AI agents. Although they work well for this domain, applying them to more complex domains may require some adjustments. Additionally, our conversation selection method groups conversations by intent. An extra step for intent classification may be needed when intent labels are unavailable. 

Our E2E evaluation aligns closely with the human judgment, making it a reliable way to assess the accuracy of workflows. However, for more complex workflows in other domains, human evaluation can provide additional insights to further refine the evaluation framework.

SynthABCD offers a controlled testing environment, ensuring consistent benchmarking. Although it does not capture the full complexity of real-world interactions, models that perform well on SynthABCD provide a good starting point. To improve robustness, we recommend additional testing on real-world datasets such as ABCD to ensure that the models perform well in diverse scenarios.


\bibliography{custom}

\appendix

\begin{figure*}[t]
\tiny 
\begin{mdframed}
\begin{minipage}{\linewidth} 
\begin{Verbatim}[commandchars=\\\{\}]
..
<think>
Okay, let's tackle this problem. The user wants me to identify the steps necessary to resolve a customer's issue based on the provided 
conversations. They specified that the steps should be essential, consolidated, and formatted as a numbered list with pre-conditions, 
required inputs, and specific conditions.

First, I need to go through all the conversations to understand the common steps agents take to resolve the return requests. Since there 
are 75 conversations, I'll look for patterns and repeated actions.

Looking at the conversations, the first thing agents usually do is ask for the customer's name or account ID to pull up their account.
Then, they request the username, email address, and order ID to validate the purchase. \textbf{Next, they check the membership level (guest,} 
\textbf{bronze, silver, gold) because the return policy varies based on that. For example, guests might have a shorter return window,}
\textbf{while gold members have unlimited returns.}

\textbf{Agents then verify the purchase date. If the purchase is beyond the standard return period for the customer's membership level, they}
\textbf{ask if the customer has the receipt or original packaging. If they do, the return is processed. Without the receipt, returns might}
\textbf{be denied, especially for guests or lower-tier members.}

After confirming eligibility, the agent collects the customer's address to generate a shipping label. Finally, they ask how the 
customer wants to process the return: by mail, in-store, or at a drop-off center.

I need to make sure all these steps are included, but without breaking them down by individual conversation. Also, pre-conditions 
like membership level affecting return eligibility and required inputs like receipt or address should be mentioned.

Let me list out the steps:

1. Collect customer's full name or account ID to access their account.
2. Obtain username, email address, and order ID to validate the purchase.
\textbf{3. Determine the customer's membership level (guest, bronze, silver, gold) to apply the appropriate return policy.}
\textbf{4. Verify the purchase date to check if it's within the allowed return period for their membership level.}
\textbf{5. If the purchase is outside the standard return period, check if the customer has the receipt or original packaging (pre-condition} 
\textbf{for returns beyond standard period).}
6. Collect the customer's shipping address to generate a return label.
7. Confirm the return method (mail, in-store, drop-off center) and provide instructions.

I should check if any steps are missing. For example, in some cases, if the customer doesn't have a receipt and it's past the return 
period, the agent can't process the return. That's covered in step 5. Also, the return method is always asked, so step 7 is necessary. 
Each step includes required inputs (like address) and conditions (membership level affecting return window).

Yes, this covers all the essential steps without unnecessary details. The list is consolidated and formatted as per the instructions.
</think>
..
\end{Verbatim}
\end{minipage}
\end{mdframed}
\caption{A snippet of <think> tokens generated by the DeepSeek-R1 model for the $return\_color$ intent using conversations from the ABCD data. Compared to the QA-CoT (Fig.~\ref{qa-cot-example}), think tokens lack fine-grained conditions based on membership level.}
\label{deepseek-think-example}
\end{figure*}

\begin{figure*}[t]
    \centering
    \includegraphics[width=0.75\linewidth]{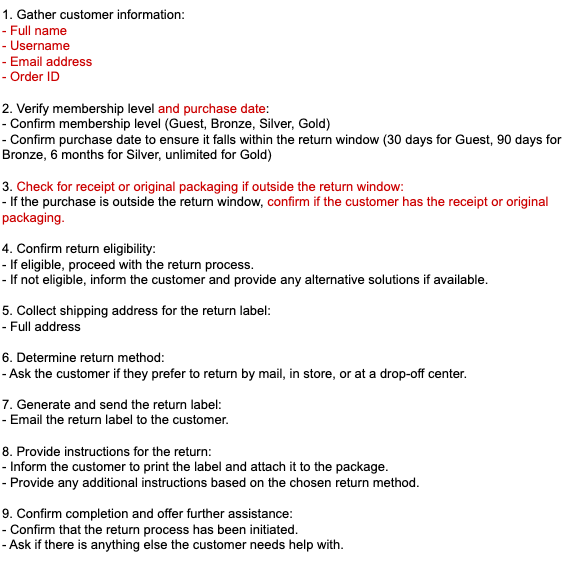}
    \caption{Example showing errors in the extracted workflow. Instead of two steps `First, ask the customer for their Full Name or Account ID to pull up account.' and `After getting the Full Name or Account ID, ask the customer for their Email Address, Order ID, and Username to validate purchase.', it combines them into one and also exclude \textbf{Account ID}. It also merge ask `membership level' and `purchase date' while the ground-truth workflow requires asking membership level first and then asking `purchase date' or `original receipt' or `original packaging' or `address' depending on the membership level.}
    \label{fig:appendix-error}
\end{figure*}

\section{Human Evaluations of E2E Framework} \label{appendix:e2e-human}
We assess the effectiveness of our E2E framework using 18 distinct intents from the ABCD dataset\footnote{All our annotators hold graduate degrees in computer science with over 8 years of experience in NLP in industry or academia.}. First, we manually evaluate the automatic scenario construction process against ground-truth workflows (Step 1, Fig.~\ref{fig:eval1}). Across these 18 intents, humans identified 231 possible scenarios, while GPT-4o generated 230, with 224 found to be correct.
Next, we evaluate the mapping of scenarios to customer information, system information, and success criteria (Steps 2 and 3, Fig.~\ref{fig:eval1}). Among the 224 correct scenarios, GPT-4o successfully generated all required details in 219 cases. Together for all three steps of the process, GPT-4o achieved an overall accuracy of 94.81\% (219 out of 231 scenarios), highlighting its effectiveness in capturing and mapping essential workflow details for user and agent bots simulation.

Finally, to evaluate the LLM's ability to predict conversation success, two annotators manually labeled successful completions for 105 simulated conversations. The manual evaluation yielded a high inter-annotator agreement, with a Cohen’s $\kappa$ of 0.966. Additionally, the agreement between the LLM's evaluation and human judgments was strong, achieving a Cohen’s $\kappa$ of 0.922. These results prove the robustness of our E2E evaluation framework in evaluating workflows.

\begin{table}[t]
\centering
\small
\begin{tabular}{lccc}
\hline
 & Macro  & Micro  &  \#utt \\ \hline
Multi-turn QA    & 52.50 & 49.28 &  9.28  \\ \hline
Single-turn QA    & \textbf{58.55} & \textbf{56.29} &  10.60  \\ \hline
\end{tabular}
\caption{Performance of multi-turn prompting and single-pass prompting for simulating the Guide-Implementer interaction. The single-pass approach achieves better accuracy.}
\label{tab:multi-step-vs-single-step-qa}
\end{table}

\section{Dataset} \label{sec:appendix-data}

\subsection{ABCD Dataset}
The original ABCD dataset includes 55 intents, but we only focus on 21 complex intents that involve multi-step workflows and discard the remaining 34, which include both non-procedural issues—such as FAQs about products, pricing, timings, membership, or features—and simpler, linear workflows. These 21 complex intents feature multiple sub-flows for handling different scenarios (e.g., different membership levels in Fig. \ref{fig:eval1}).


\subsection{SynthABCD Dataset} \label{appendix:synthabcd}
Among the 21 intents used in our E2E study, three workflows are duplicates of each other, and two others are also duplicates, resulting in a total of 18 distinct workflows. We focus on these 18 intents. To introduce variability, we create 50 random user profiles with varying attributes such as name, profession, and city of residence, generating two distinct synthetic conversations for each scenario.
The prompts used for converting workflows into scenarios and for simulating conversations are provided in Fig. \ref{prompt:synthabcd-e2e-step1} and Fig. \ref{prompt:synthabcd-conv}, respectively.

Our SynthABCD pipeline employs the same automatic scenario construction process as the E2E framework. Across the 18 intents, human annotation identified 231 possible scenarios, while the GPT-4o model generated 230 scenarios. Of these, 224 were correctly extracted, yielding an accuracy of 96.97\%. This high accuracy—with only one scenario missed and six incorrect extractions—demonstrates the effectiveness of our conversation synthesis process and its applicability to new workflows.
All intents used in our evaluation are reported in Table \ref{table:appendix-data-stats}.

\begin{table*}[t]
\centering
\small
\begin{tabular}{l|rccc} \hline
Intent & Type & \#Secnarios & ABCD & SynthABCD  \\ \hline
  refund\_initiate &  simple & 1 & & \\
  refund\_update &  simple & 1 & & \\
  refund\_status &  simple & 1 &  & \\
  return\_stain & complex & 18 & $\checkmark$ & $\checkmark$   \\
  return\_size & complex & 18 & $\checkmark$ &    \\
  return\_color & complex & 18 & $\checkmark$ &    \\
  status\_mystery\_fee & complex & 10 & $\checkmark$ & $\checkmark$    \\
  status\_delivery\_time & complex & 10 & $\checkmark$ & $\checkmark$  \\
  status\_payment\_method & complex & 8 & $\checkmark$ & $\checkmark$   \\
  status\_quantity & complex  & 8 & $\checkmark$ & $\checkmark$   \\
  manage\_upgrade & complex  & 14 & $\checkmark$ & $\checkmark$   \\
  manage\_downgrade & complex & 26 & $\checkmark$ & $\checkmark$  \\
  manage\_create & complex &  26 &$\checkmark$ & $\checkmark$  \\
  manage\_cancel & complex & 26 &  $\checkmark$ & $\checkmark$ \\
  recover\_username &  simple & 1 & &  \\
  recover\_password &  simple & 1 & &  \\
  reset\_2fa &  simple & 1 & &  \\
  status\_service\_added & complex  & 10 & $\checkmark$ & $\checkmark$    \\
  status\_service\_removed & complex   & 8 & $\checkmark$ & $\checkmark$    \\
  status\_credit\_missing & complex   & 6 &  $\checkmark$ & $\checkmark$    \\
  manage\_change\_address &  simple & 1 & &  \\
  manage\_change\_name &  simple & 1  & &  \\
  manage\_change\_phone &  simple & 1  & & \\
  manage\_payment\_method &  simple & 1  & &  \\
  bad\_price\_competitor &  simple & 1  & & \\
  bad\_price\_yesterday &  simple & 1  & &   \\
  out\_of\_stock\_general &  simple  & 1 & &   \\
  out\_of\_stock\_one\_item &  simple  & 1 & &   \\
  promo\_code\_out\_of\_date & complex   & 12 & $\checkmark$ & $\checkmark$    \\
  promo\_code\_invalid & complex  & 12 & $\checkmark$ & \\
  mistimed\_billing\_already\_returned & complex  & 10 & $\checkmark$ & $\checkmark$   \\
  mistimed\_billing\_never\_bought & complex  & 10 & $\checkmark$ & $\checkmark$  \\
  missing &  simple  & 1 & & \\
  cost & complex  & 8 &  $\checkmark$ & $\checkmark$    \\
  status\_due\_amount &  simple  & 1&  &   \\
  status\_due\_date &  simple  & 1&  &   \\
  manage\_pay\_bill &  simple  & 1&  &  \\
  manage\_extension & complex   & 8 & $\checkmark$ & $\checkmark$    \\
  manage\_dispute\_bill & complex   & 12 & $\checkmark$ & $\checkmark$  \\
  credit\_card &  simple  & 1& &  \\
  shopping\_cart &  simple   & 1& &   \\
  search\_results &  simple   & 1& &   \\
  slow\_speed &  simple   & 1& &  \\
  \hline
\end{tabular}
\caption{Intents used in E2E evaluation framework for the ABCD and SynthABCD datasets. return\_stain, return\_size and return\_color share the same workflow, as do promo\_code\_out\_of\_date and promo\_code\_invalid. Therefore, we only use one from each group in SynthABCD.} \label{table:appendix-data-stats}
\end{table*}

\subsection{Data Quality} \label{sec:data-compliance}
\begin{table}[t]
\centering
\small
\begin{tabular}{lcccc}
\hline
\textbf{Dataset} & \textbf{F} & \textbf{NA} & \textbf{NF} & \textbf{NC Cnv.}  \\ \hline
ABCD & 39.0\% & 57.07\% & 3.93\% & 28.57\% \\
SynthABCD & 43.56\% & 56.21\% & 0.22\% & 2.71\% \\ \hline
\end{tabular}
\caption{Data quality evaluation of ABCD and SynthABCD datasets. F: followed, NA: not applicable, NF: not followed, NC Cnv.: non-compliant conversation}
\label{tab:data-quality}
\end{table}

We evaluate the compliance of conversations in both the ABCD validation set and SynthABCD with their corresponding ground-truth workflows using the GPT-4o model (Prompt in Fig. \ref{prompt:compliance}). For each conversation, we assess whether each step in the workflow is followed, deemed not applicable, or not followed. Additionally, we calculate the percentage of non-compliant conversations, which refers to the conversations that fail to comply with at least one step in the ground-truth workflows. 

The data quality evaluation results in Table \ref{tab:data-quality} show key differences between the ABCD and SynthABCD datasets. In ABCD, 39.0\% of workflow steps are followed, compared to 43.56\% in SynthABCD.  Additionally, the percentage of unfollowed steps (NF) is significantly lower in SynthABCD (0.22\%) compared to ABCD (3.93\%), demonstrating the effectiveness of controlled conversation synthesis using the LLM. This reduction in noise aligns with SynthABCD’s objective of enhancing data consistency. Consequently, SynthABCD achieves 97.29\% compliant conversations, a substantial improvement over ABCD’s 71.43\%, reflecting its higher consistency.

\begin{figure}[]
    \centering
    \includegraphics[width=0.99\linewidth]{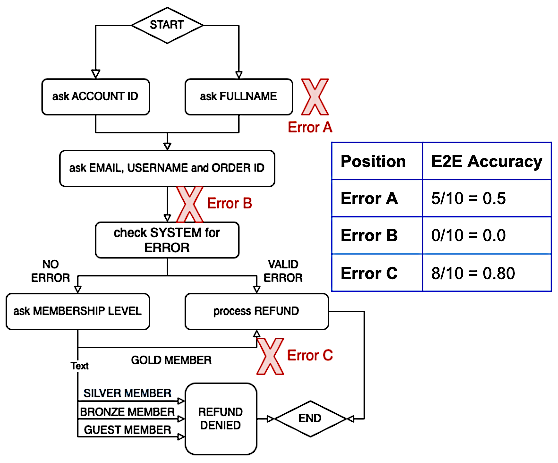}
    \caption{Illustration of End-to-End (E2E) accuracy evaluation. For this workflow, there are 10 scenarios based on membership level, system error, and whether the user can provide their full name or account ID. In the case of error B, none of the scenarios can be completed successfully, resulting in an accuracy of 0. For error A, only 50\% of scenarios where the user provides an account ID can be completed, yielding an accuracy of 0.5. For error C, 2 scenarios fail, including those for gold members with either account ID or full name, resulting in an accuracy of 0.8.}
    \label{fig:eval-e2e}
\end{figure}

\section{Alternative Evaluation Frameworks for Dialog Workflows} \label{sec:alternative-evaluation}
In addition to our primary End-to-End (E2E) evaluation, we explored several alternative evaluation frameworks to assess step-level correctness in workflow reconstruction. While these approaches provide fine-grained insights, they fail to capture the holistic effectiveness of a workflow and exhibit low agreement with human evaluations. Notably, methods that evaluate steps in isolation overlook dependencies across the entire process, potentially leading to an over- or under-estimation of practical performance. As illustrated in Fig.~\ref{fig:eval-e2e}, not all errors have the same impact on workflow completion—some may block all scenarios (e.g., error B with 0\% accuracy), while others only affect a subset of cases (e.g., error C reducing accuracy to 0.8). Step-level evaluation treats all errors as equal, failing to reflect their true effect on workflow execution. In contrast, E2E evaluation remains the most reliable metric, as it closely aligns with human judgments and directly measures real-world utility. Since workflows are ultimately assessed based on their ability to guide complete processes, we adopt E2E evaluation as our primary strategy.

\subsection{Evaluation Methods}

\subsubsection{Automatic QA Evaluator} 

We explored a QA-based evaluation framework that measures the accuracy of each step in a workflow by verifying its correct dependence on preconditions, such as the outcome of a previous step \cite{fabbri2021qafacteval}. For example, if the ground truth workflow includes a step to ask shipment status' followed by either ask membership level if the product has shipped' or `issue refund if the product has not shipped,' our goal is to verify that the predicted workflow reflects this conditional logic accurately.
To achieve this, we decompose ground-truth workflows into atomic questions, such as:
Q: What is the next step after asking shipment status if the product has not shipped? A: issue refund.
Q: What is the next step after asking shipment status if the product has shipped? A: ask membership level.
By structuring evaluation in this way, we ensure a step-wise correctness check while maintaining consistency. We use the GPT-4o model both for generating QA pairs from ground-truth workflows and for evaluating the predicted workflows. 

\subsubsection{Embedding Similarity}
Embedding similarity method quantifies structural and semantic similarity using cosine distance between workflow representations. Specifically, we represent both the ground-truth and predicted workflows as embedding vectors and define the similarity score as:  

\begin{equation}
S_{\text{embedding}} = \frac{1}{\text{cosine\_distance}(\mathbf{w}_{\text{ref}}, \mathbf{w}_{\text{pred}})}
\end{equation}

where \( \mathbf{w}_{\text{ref}} \) and \( \mathbf{w}_{\text{pred}} \) are the vector representations of the ground-truth and predicted workflows, respectively. A higher score indicates greater alignment between the two workflows.  
To obtain these embeddings, we use OpenAI text-embedding-3-small model.

\subsubsection{Edit Distance}
Edit-distance-based evaluation quantifies structural differences by computing the minimum number of operations required to transform the predicted workflow into the ground-truth workflow. We define three types of operations:  

\begin{itemize}
    \item \textbf{Insertion}: Adding a missing step to align with the ground-truth workflow.
    \item \textbf{Deletion}: Removing an extraneous step that does not appear in the ground-truth workflow.
    \item \textbf{Reordering}: Adjusting the sequence of steps to match the ground-truth workflow.
\end{itemize}  

To compute edit distance, we use the GPT-4o that aligns the predicted workflow to the ground-truth workflow using these operations. The final score is defined as:  

\begin{equation}
S_{\text{edit}} = \frac{1}{\text{number\_of\_edits}}
\end{equation}

where a higher score indicates fewer modifications needed for alignment, implying better prediction accuracy.

\subsubsection{Step-Accuracy Evaluation}  
We use the GPT-4o model to label each step in the ground-truth workflow as either correctly or incorrectly covered with respect to the predicted workflow. The final score is computed as:  

\begin{equation}
S_{\text{step}} = \frac{\text{number\_of\_correct\_steps}}{\text{total\_number\_of\_steps}}
\end{equation}

where a higher score indicates that a greater proportion of steps in the ground-truth workflow are correctly captured by the predicted workflow.

\subsubsection{Likert-Scale Evaluation}  
We use the GPT-4o model to assign a similarity score between 1 and 100 based on structural and semantic alignment between predicted and ground-truth workflows. The score is computed as:  

\subsection{Human Evaluation Setup}
To evaluate the alignment of alternative evaluation protocols, we conduct a human study on workflows for 11 complex intents, each accompanied by 4 predicted workflows. The evaluation process began with decomposing each workflow into a series of QA pairs. For the gold ground-truth workflows, one author wrote the QA pairs, while a second author verified their quality. This dual-step validation ensures that the questions were clear and accurately represented the underlying steps and conditional logic inherent in the workflows. From the 11 workflows, we obtain a total of 141 QA pairs. Additionally, the 4 predicted workflows for each intent collectively yield 564 QA pairs, providing a comprehensive dataset for evaluation.

Next, two authors independently labeled the correctness of each QA pair based on the corresponding predicted workflow. This independent labeling process resulted in a Cohen's $\kappa$ score of 0.742, indicating substantial agreement between the annotators. 
For QA pairs labeled as correct and incorrect, we achieve an F1 score of 0.937 for correct labels, demonstrating high agreement on identifying correct steps, and 0.82 for incorrect labels, indicating slightly lower agreement in identifying errors. This further indicates that detecting errors by reasoning on previous steps and their outcomes is challenging, even for humans.

To assess the overall alignment between the predicted and ground-truth workflows, we calculate a final score for each predicted workflow based on the percentage of correctly answered questions. This score shows a correlation score of 0.835 between human annotators, indicating strong consistency in their evaluations. 
We further analyze correlations between step-level evaluation methods and human judgments, as reported in Table~\ref{tab:evaluator-correlation}. Among automated evaluation methods, the QA-based metric achieves the highest correlation (0.708) with human annotations, indicating its relative effectiveness in step-level evaluation. However, these correlation scores remain lower than those observed with End-to-End (E2E) evaluation, reinforcing our decision to prioritize E2E evaluation as the primary metric.  

\begin{table}[t]
\centering
\small
\begin{tabular}{lcc}
\hline
\textbf{Automatic Evaluation} & \textbf{Correlation} \\ \hline
embedding             & 0.412                \\
edit-distance               & 0.587                \\
step-accuracy              & 0.631                \\
likert-scale             & 0.678                \\
QA-based (Proposed)                 & \textbf{0.708 }               \\ \hline
\end{tabular}
\caption{Correlation between different automatic evaluation metrics and human judgements.}
\label{tab:evaluator-correlation}
\end{table}

\section{Prompts}

\begin{figure*}[!h]
\tiny
\begin{mdframed}
Extract intent, slot values and resolution steps from the customer service chat provided under <CONVERSATION> section below.

STEP-BY-STEP INSTRUCTIONS

1. Exclude any non-essential statements such as greetings, apologies, and expressions of gratitude.

2. Be concise, complete, and truthful to the facts mentioned in the conversation.

3. Include any all the numberical and categorical details such as personal names, addresses, phone numbers etc.

4. Follow the exact sequential order in which the agent took different actions in "resolution\_step".

5. Respond using the JSON format:

\begin{verbatim}
{
    "intent": "What is the customer issue? Use max 50 characters.",
    "slot_values": {
        "dictionary of slots and values collected from the customer or provided to the customer by the agent formatted as {slot: value}"
    },
    "resolution_steps": ["List of steps/ actions taken by the agent to resolve the customer's issue."]
}
\end{verbatim}

Example Conversation:

Customer: Hi, I'd like to return an item I purchased.

Agent: Hi! I’m happy to help with that. Could you please provide me with your order ID so I can pull up your account?

Customer: Sure, it’s 123456.

Agent: Thank you. Can you confirm the reason for your return?

Customer: The item arrived damaged.

Agent: I’m sorry to hear that. We’ll get this sorted for you right away. Would you prefer a replacement or a refund?

Customer: I’d like a refund, please.

Agent: Got it. I’ve initiated the return process for a refund. You’ll receive an email with a prepaid return label.Once we receive the item, we’ll process your refund within 3–5 business days. Can I help you with anything else?

Customer: No, that’s all. Thanks!

Agent: You’re welcome! If you need further assistance, feel free to reach out. Have a great day!

Example Output:
\begin{verbatim}
{
    "intent": "Customer wants to return a damaged item.",
    "slot_values": {
        "order_id": "123456",
        "return_reason": "damaged item",
        "refund_preference": "refund"
    },
    "resolution_steps": [
        "Agent asked for the order ID.",
        "Agent asked the reason for the return.",
        "Agent asked whether the customer wants a replacement or refund.",
        "Agent initiated the return process for a refund.",
        "Agent informed the customer about receiving a prepaid return label via email.",
        "Agent explained that the refund will be processed within 3–5 business days after receiving the item."
    ]
}
\end{verbatim}
\end{mdframed}
\caption{The prompt for extracting key elements: issue description, slots-values and resolution steps from ABCD conversations [Conversation Selection].} \label{prompt:conv-selection-procedure}
\end{figure*}

\begin{figure*}[!h]
\tiny
\begin{mdframed}
Identify all branching conditions in the provided policy. Do not include common steps. Generate one branch per line.

Example Policy 1:

\{policy\_1\}

Example Output 1:

\{output\_1\}

Example Policy 2:

\{policy\_2\}

Example Output 2:

\{output\_2\}

\end{mdframed}
\caption{The prompt for mapping workflows to scenarios [SynthABCD and E2E Evaluation].} \label{prompt:synthabcd-e2e-step1}
\end{figure*}

\begin{figure*}[!h]
\tiny
\begin{mdframed}
You are given below a dialog policy that a customer service agent uses to resolve a customer's issue.

\#\# Policy

{policy}

Read the policy carefully and simulate a conversation between an user and an agent that follows the below sub-policy.

\{subflow\}

\# Instruction

- Generate conversation that strictly follow the provided sub-policy.

- Make sure that the conversation is started by user. 

User's details

Name: \{user\_name\}

Profession: \{user\_profession\}

Address: \{city\}

- Make sure that agent only asks the information in subflow or common steps in the dialog policy. Avoid any unnecessary information.

Example Policy 1:

\{example\_policy\}

Example Subflow 1:

\{example\_subflow\}

Example Output 1:

\{example\_conv\}

\end{mdframed}
\caption{The prompt for generating conversations given a workflow and a scenario [SynthABCD].} \label{prompt:synthabcd-conv}
\end{figure*}

\begin{figure*}[!h]
\tiny
\begin{mdframed}
You are given a dialog workflow and a specific user scenario. Generate detailed list of user information, system information and the corresponding expected outcome for the given scenario based on the workflow. Respond following the below json format.

\begin{verbatim}
{
    "user information": ["detailed list of information that the user provide to the agent to complete the conversation in the given scenario"],
    "system information": ["detailed list of information that the agent check in system to complete the conversation in the given scenario"],
    "outcome": "expected outcome reflecting the resolution in the given scenario"
}
\end{verbatim}

Example Policy 1:

\{policy\_1\}

Example Scenario 1:

\{scenario\_1\}

Example Output 1:

\{output\_1\}

Example Policy 2:

\{policy\_2\}

Example Scenario 2:

\{scenario\_2\}

Example Output 2:

\{output\_2\}

Example Policy 3:

\{policy\_3\}

Example Scenario 3:

\{scenario\_3\}

Example Output 3:

\{output\_3\}

Example Policy 4:

\{policy\_4\}

Example Scenario 4:

\{scenario\_4\}

Example Output 4:

\{output\_4\}

\end{mdframed}
\caption{The prompt for extracting user information, system information and success criteria given a scenario and a workflow [E2E evaluation].} \label{prompt:e2e-step2}
\end{figure*}

\begin{figure*}[!h]
\tiny
\begin{mdframed}
You are given a dialog policy and corresponding criteria for successful completion of conversation with a customer.

Given a conversation between an agent and a customer, check if the conversation ends successfully or not.

Dialog Policy:

\{policy\}

Success Criteria:

\{outcome\}

Conversation:

\{conv\}

Respond in json following the format below.

\begin{verbatim}
{
    "successful": "yes/no"
    "explanation": "explain how the conversation went?"
}
\end{verbatim}
\end{mdframed}
\caption{The prompt for evaluating successful completion of a conversation between the customer and service bot [E2E evaluation].} \label{prompt:e2e-success}
\end{figure*}

\begin{figure*}[!h]
\tiny
\begin{mdframed}
You are a customer talking to an agent to have your issue resolved. You are given the issue description and the slots-values that you share with the agent. Respond to the agent and provide them them the requested information if possible.

STEP-BY-STEP INSTRUCTIONS:

- Identify the information requested by the agent.

- Check the provided issue description and the slots-values and identify if you can provide the requested information to the agent.

- If you can, respond to the agent with the requested information.

- If you don't have the requested information, respond that "you don't have the requested information and ask if there is any other information you can provide".

- Only provide the information that is requested by the user.

\#\# Issue:

\{issue\}

\#\# Slots-Values that you can provide to the agent:

{info}

Make up free form slot-values, e.g., user name, full name, order ID, account ID, email and address. 

\#\# Conversation History:

\{history\}
\end{mdframed}
\caption{The prompt for simulating a customer bot [E2E Evaluation].} \label{prompt:customer-bot-e2e}
\end{figure*}

\begin{figure*}[!h]
\tiny
\begin{mdframed}
You are a customer service agent trying to solve a customer's issue. You are given the dialog policy, conversation history and the system information. Respond to the customer by strictly following the provided dialog policy.

STEP-BY-STEP INSTRUCTIONS:

- Read the provided conversation history and identify the current dialog state.

- Match dialog state with the dialog policy, and identify the next step to address the customer's issue.

- If the next step involves checking system information (e.g., checking system for an error), check the available system information and inform that to the customer. If the required system information is not available, reply "DONE".

- If the next step involves requesting information from customer (e.g., username), ask customer for the required information.

- If the next step involves taking some action (e.g., issuing refund), inform the customer that you have succesfully taken that action.

- Respond to the customer based on the identified next step.

- If the policy does not describe the next step based on the current dialog state, conclude the conversation by generating "DONE".

- If you can not take the next step for any reason, conclude the conversation by generating "DONE".

- If the issue has been successfully resolved, conclude the conversation by generating "DONE".

- Avoid repeating yourself or requesting information that has already been mentioned in the conversation history.

\#\# Dialog Policy

```

\{policy\}

```

\#\# System Information

\{info\}

\#\# Conversation History

\{history\}

\end{mdframed}
\caption{The prompt for simulating a service agent bot [E2E Evaluation].} \label{prompt:service-bot-e2e}
\end{figure*}

\begin{figure*}[!h]
\tiny
\begin{mdframed}
You are a quality assurance manager tasked to assess whether an agent followed the established agent guidelines for resolving a customer's issue. The guidelines offer detailed, rule-based instructions for agents to be followed in a step-by-step manner. However, agents may not consistently adhere to these instructions. Your task is to evaluate the agent's compliance by comparing the steps taken in the provided conversation against the established guidelines.

**Important: Approach your task step-by-step. Carefully evaluate each rule in the guidelines.**

\#\# Step-by-Step Instruction

- For each rule in guidelines, determine whether that rule is applicable to conversation or not.

  - Some rules are context-specific, they define actions contingent on preceding action outcomes, making them applicable only in specific conversational situations.
  
  - Other rules are universally applicable, dictating actions that consistently follow a preceding action, regardless of the outcome.

- If the rule is applicable, check whether the agent followed the prescribed steps accurately or not.

  - Evaluate the accuracy of both the action, and the order of the action.

- Here are some rule for handling specific scenarios

  - If the customer has already provided certain details (such as Full Name, Account ID, refund amount, item details, etc.) either voluntarily or as requested by the agent in a prior step:
  
    - Do not penalize the agent for not following guidelines by not requesting the same details again. 
    
    - Do not penalize the agent for not following guidelines even if the agent request for the same details again.
    
  - If a rule includes multiple actions, the order of those actions is irrelevant and should not be penalized. For instance, [ask 'Email Address', 'Order ID', and 'Username'] or  [ask 'Order ID', 'Email Address', and 'Username'] are equivalent.

- Generate output in the JSON format given below, offering both the response and an explanation for each rule.

\begin{verbatim}
{
    "Rule_1":{
    "response": 'followed'/ 'not applicable'/ 'not followed',
    "explanation": ''
    },
    "Rule_2":{
    "response": 'followed'/ 'not applicable'/ 'not followed',
    "explanation": ''
    },
    ..
}
\end{verbatim}

\end{mdframed}
\caption{The prompt for conversation's compliance evaluation given the ground-truth workflow [Data Quality].} \label{prompt:compliance}
\end{figure*}









\begin{figure*}[!h]
\tiny
\begin{mdframed}
You are a QA simulator consisting of two agents: a Guide and an Implementer. Both agents are tasked with collaboratively reconstructing the process of resolving customer issues by analyzing historical conversations.

Below are the profiles for Guide and Implementer:

\#\#\# Guide:

- Asks targeted questions to understand the steps taken in a given scenario.

- Focuses on clarifying the preconditions, decision points, and the logic behind the steps.

- Ensures that all possible customer contexts and edge cases are addressed.

\#\#\# Implementer:

- Review past conversations to answer the questions.

- Provides detailed explanations about the actions taken at each step and their rationale.

\#\#\# Below is an example discussion between a guide and an implementer agent:

Guide: What is the first step an agent should take when booking a restaurant?

Implementer: ask name and phone number

Guide: What is the next step after asking the name and phone number? 

Implementer: ask preferred date and time for the reservation

Guide: What is the next step after asking the preferred date and time for the reservation? 

Implementer: ask party size

Guide: What is the next step after asking the party size? 

Implementer: ask for any special requests

Guide: What is the next step after asking about any special request? 

Implementer: check availability

Guide: What is the next step if restaurant is booked at requested date and time? 

Implementer: offer alternative time and date

Guide: What is the next step if restaurant is available at requested date and time?

Implementer: book and inform the customer

Let's think step-by-step and generate a discussion between the Guide and the Implementer based on below conversations.
\end{mdframed}
\caption{The prompt for generating question-answer-based chain-of-thoughts [Workflow Extraction: QA-CoT].} \label{prompt:qa-cot}
\end{figure*}

\begin{figure*}[!h]
\tiny
\begin{mdframed}
Identify the steps necessary to resolve the customer's issue based on the provided conversations and the discussion between Guide and Implementer Agents.

- Include only the essential actions needed for issue resolution, excluding unnecessary steps.

- Use the discussion between Guide and Implementer Agents to identify important steps in the workflow.

- Create a unified, consolidated list of steps without breaking them down by individual conversation.

- Formatting Instructions:

    1. Use simple and concise language for each step, mentioning any pre-conditions where applicable.
    
    2. Organize the steps in a numbered list for clarity.
    
    3. Include relevant details, such as required inputs and specific conditions, for each step.
\end{mdframed}
\caption{The prompt for extracting workflows given historical conversations and the discussion between the guide and implementer agents [Workflow Extraction: QA-CoT].} \label{prompt:procedural-extraction}
\end{figure*}

\begin{figure*}[!h]
\tiny
\begin{mdframed}
Identify the steps necessary to resolve the customer's issue based on the provided conversations.

- Include only the essential actions needed for issue resolution, excluding unnecessary steps.

- Create a unified, consolidated list of steps without breaking them down by individual conversation.

- Formatting Instructions:

    1. Use simple and concise language for each step, mentioning any pre-conditions where applicable.
    
    2. Organize the steps in a numbered list for clarity.
    
    3. Include relevant details, such as required inputs and specific conditions, for each step.
\end{mdframed}
\caption{The prompt for basic prompting strategy. [Workflow Extraction Baseline: Basic].} \label{prompt:procedural-extraction-basic}
\end{figure*}

\begin{figure*}[!h]
\tiny
\begin{mdframed}
Reflect on the following past chats and assess how well the dialog workflow covers the different scenarios mentioned. Your goal is to focus on the most representative scenarios discussed in the chats, excluding outlier conversations where agent may have taken wrong steps. For each valid scenario, consider whether the workflow addresses it comprehensively, and if not, identify the gaps.
        
Evaluate Coverage Gaps

    - Where does the workflow fail to provide guidance or actions for specific scenarios presented in the past chats?
    
    - Are there any special cases (e.g., different membership tiers, exceptions, or advanced user queries) that the workflow misses entirely?
    
    - Ignore outlier conversations where agent actions are inconsistent with most other conversations from the same scenario.
    
    - For each gap identified, explain how the current workflow might be extended or modified to address these uncovered cases.
    
Suggestions for Enhancing Coverage

    - What additions or modifications could be made to the workflow to cover the uncovered scenarios more effectively?

\end{mdframed}
\caption{The prompt for identifying coverage gap in a workflow given historical conversations. [Workflow Extraction Baseline: Reflect].} \label{prompt:procedural-extraction-reflect}
\end{figure*}

\begin{figure*}[!h]
\tiny
\begin{mdframed}
Improve the below dialog workflow based on the provided conversations and the LLM feedback.

- Include only the essential actions needed for issue resolution, excluding unnecessary steps.

- Create a unified, consolidated list of steps without breaking them down by individual conversation.

- Ignore feedback that is specific to an individual customer and does not address a general dialog scenario.

- Formatting Instructions:

    1. Use simple and concise language for each step, mentioning any pre-conditions where applicable.
    
    2. Organize the steps in a numbered list for clarity.
    
    3. Include relevant details, such as required inputs and specific conditions, for each step.
\end{mdframed}
\caption{The prompt for improving coverage given conversations and LLM-feedback. [Workflow Extraction Baseline: Reflect].} \label{prompt:procedural-extraction-reflect-generate}
\end{figure*}

\begin{figure*}[!h]
\tiny
\begin{mdframed}
Your task is to identify the steps necessary to resolve the customer's issue based on the provided conversations. 

\{convs\}

Let's first understand the problem and devise a plan to solve the problem.
\end{mdframed}
\caption{The prompt for generating plan from historical conversations. [Workflow Extraction Baseline: Plan].} \label{prompt:procedural-extraction-plan}
\end{figure*}

\begin{figure*}[!h]
\tiny
\begin{mdframed}
Identify the steps necessary to resolve the customer's issue based on the provided conversations and the plan.

- Include only the essential actions needed for issue resolution, excluding unnecessary steps.

- Use the plan to identify important steps in the workflow.

- Create a unified, consolidated list of steps without breaking them down by individual conversation.

- Formatting Instructions:

    1. Use simple and concise language for each step, mentioning any pre-conditions where applicable.
    
    2. Organize the steps in a numbered list for clarity.
    
    3. Include relevant details, such as required inputs and specific conditions, for each step.
    
- Only generate the final guideline, outlining important steps required for solving customer's issue.
\end{mdframed}
\caption{The prompt for generating workflows from plan and historical conversations. [Workflow Extraction Baseline: Plan].} \label{prompt:procedural-extraction-plan-generating}
\end{figure*}

\begin{figure*}[!h]
\tiny
\begin{mdframed}
Aggregate and improve upon the following four dialog workflows used to resolve the customer's issue. 

Workflow 1:

\{wf1\}

Workflow 2:

\{wf2\}

Workflow 3:

\{wf3\}

Workflow 4:

\{wf4\}

Use the provided example conversations to decide which of the mentioned workflows is correct at each step or to identify any steps that are missing from all workflows.

\{convs\}

STEP-BY-STEP INSTRUCTIONS:

- Identify all the steps necessary to resolve the customer's issue.

- Include only the essential actions needed for issue resolution, excluding unnecessary steps.

- Create a unified, consolidated list of steps without breaking them down by individual conversation.

- Formatting Instructions:

    1. Use simple and concise language for each step, mentioning any pre-conditions where applicable.
    
    2. Organize the steps in a numbered list for clarity.
    
    3. Include relevant details, such as required inputs and specific conditions, for each step.
\end{mdframed}
\caption{The prompt for ensemble prompting strategy. [Workflow Extraction Baseline: Ensemble].} \label{prompt:procedural-extraction-ensemble}
\end{figure*}

\begin{figure*}[!h]
\tiny
\begin{mdframed}
Correct the given LLM-generated discussion between two collaborative agents, the Guide and the Implementer. 

Reflect on the given discussion:

    - Evaluate whether the discussion cover all the steps comprehensively.

    - Look for discussion parts that may belong to outlier conversations (e.g., rare scenarios or edge cases) and determine if they should be part of the main workflow.

Correct and Complete:

    - Add any missing discussions to bridge gaps and ensure logical continuity.
    
    - Reorganize or exclude discussion about steps in outlier conversations if they do not fit the main workflow.
    
    - Clearly specify the conditions under which any alternative steps or edge cases arise.
    
Provide the corrected and complete discussion following the format of original discussion between the guide and implementer agents. Do not add any justifications.
\end{mdframed}
\caption{The prompt used for refining the QA chain-of-thought [Workflow Extraction Baseline: QA-CoT+Reflect].} \label{prompt:procedural-extraction-qa-reflect}
\end{figure*}

\end{document}